\documentclass[10pt,twocolumn,letterpaper]{article}

\usepackage{cvpr}
\usepackage{times}
\usepackage{epsfig}
\usepackage{graphicx}
\usepackage{amsmath}
\usepackage{amssymb}
\usepackage{ifthen}
\usepackage{fancyhdr}
\graphicspath{ {images/} }

\usepackage[pagebackref=true,breaklinks=true,colorlinks,bookmarks=false]{hyperref}

\cvprfinalcopy 

\makeatletter
\newcommand*\titleheader[1]{\gdef\@titleheader{#1}}
\AtBeginDocument{
	\let\st@red@title\@title
	\def\@title{%
		\bgroup\normalfont\large\centering\@titleheader\par\egroup
		\vskip1.5em\st@red@title}
}
\makeatother

\title{Eye in the Sky: Real-time Drone Surveillance System (DSS) for Violent Individuals Identification using ScatterNet Hybrid Deep Learning Network}
\titleheader{To Appear in the IEEE Computer Vision and Pattern Recognition (CVPR) Workshops 2018} 

\begin{document}

\author{Amarjot Singh\\
Department of Engineering\\
University of Cambridge, U.K.\\
{\tt\small as2436@cam.ac.uk} \\
\and
Devendra Patil\\
National Institute of Technology\\
Warangal, India\\
{\tt\small pdevendra@student.nitw.ac.in} \\
\and
SN Omkar\\
Indian Institute of Science\\
Bangalore, India\\
{\tt\small omkar@aero.iisc.ernet.in} \\
}

\maketitle

\begin{abstract}
Drone systems have been deployed by various law enforcement agencies to monitor hostiles, spy on foreign drug cartels, conduct border control operations, etc. This paper introduces a real-time drone surveillance system to identify violent individuals in public areas. The system first uses the Feature Pyramid Network to detect humans from aerial images. The image region with the human is used by the proposed ScatterNet Hybrid Deep Learning (SHDL) network for human pose estimation. The orientations between the limbs of the estimated pose are next used to identify the violent individuals. The proposed deep network can learn meaningful representations quickly using ScatterNet and structural priors with relatively fewer labeled examples. The system detects the violent individuals in real-time by processing the drone images in the cloud. This research also introduces the aerial violent individual dataset used for training the deep network which hopefully may encourage researchers interested in using deep learning for aerial surveillance. The pose estimation and violent individuals identification performance is compared with the state-of-the-art techniques.
\end{abstract}

\section{Introduction}
The rate of criminal activities by individuals and threats by terrorist groups has been on the rise in recent years. The law enforcement agencies have been motivated to use video surveillance systems to monitor and curb these threats. Many automated video surveillance systems have been developed in the past to monitor abandoned objects (bags)~\cite{li2010abandoned}, theft~\cite{chuang2009carried}, fire or smoke~\cite{seebamrungsat2014fire}, violent activities~\cite{goya2009method}, etc.

Li et al.~\cite{li2010abandoned} developed a video surveillance system to identify the abandoned objects with the use of Gaussian mixture models and Support Vector Machine. This system is robust to illumination changes and performs with an accuracy of 84.44\%. This system is vital for the detection of abandon bags in busy public areas, which may contain bombs. Chuang et al.~\cite{chuang2009carried} used Forward-backward ratio histogram and a finite state machine to recognize robberies. This system has proven to be very useful around automatic teller machines (ATMs) and has detected 96\% cases of the theft. Seebamrungsat et al.~\cite{seebamrungsat2014fire} presented a fire detection system based on HSV and YCbCr color models as it allowed it to distinguish bright images more efficiently than other RGB models. The system has been shown to detect fire with an accuracy of more than 90.0\%. Goya et al.~\cite{goya2009method} introduced a Public Safety System (PSS) for identifying criminal actions such as purse snatching, child kidnapping, and fighting using distance, velocity, and area to determine the human behavior. This system can identify the criminal actions with an accuracy of around 85\%.

These reported systems have been very successful in detecting and reporting various criminal activities. Despite their impressive performance (more than 90\% accuracy), the area these systems can monitor is limited due to the restricted field of view of the cameras. The law enforcement agencies have been motivated to use aerial surveillance systems to surveil large areas. Governments have recently deployed drones in war zones to monitor hostiles, to spy on foreign drug cartels~\cite{ padgett2009drones}, conducting border control operations~\cite{ walters2010ucav} as well as finding criminal activity in urban and rural areas~\cite{ lewis2010cctv}. One or more soldiers pilot most of these drones for long durations which makes these systems prone to mistakes due to the human fatigue. 

Surya et al.~\cite{penmetsa2014autonomous} proposed an autonomous drone surveillance system capable of detecting individuals engaged in violent activities in public areas. This first of its kind system used the deformable parts model~\cite{felzenszwalb2005pictorial} to estimate human poses which are then used to identify the suspicious individuals. This is an extremely challenging task as the images or videos recorded by the drone can suffer from illumination changes, shadows, poor resolution, and blurring. Also, the humans can appear at different locations, orientations, and scales. Despite the above-explained complications, the system can detect violent activated with an accuracy of around 76\% which is far less as compared to the greater than 90\% performance of the ground surveillance systems.

This paper introduces an improved real-time autonomous drone surveillance system to identify violent individuals in public areas. The proposed method first uses the feature pyramid network (FPN)~\cite{hd} is used to detect the humans from the aerial image. Next, the proposed ScatterNet Hybrid Deep Learning (SHDL) network is used to estimate the pose for each detected human. Finally, the orientations between the limbs of the estimated pose are used by the support vector machine (SVM) to identify individuals engaged in violent activities. 

The novelties of the proposed system and the advantages over Surya et al.'s~\cite{penmetsa2014autonomous} technique are detailed below: 

\begin{itemize}
	\setlength\itemsep{-0.35em}
\item \textbf{\textit{Accurate Human Pose Estimation}}: The proposed system uses the SHDL for human pose estimation. Deep networks have achieved the state-of-the-art pose estimation performance with high-level features~\cite{li20143d, pfister2014deep, toshev2014deeppose} which gives the proposed system a competitive edge.

\begin{figure}[t]
	\begin{center}
		\includegraphics[width=1\linewidth]{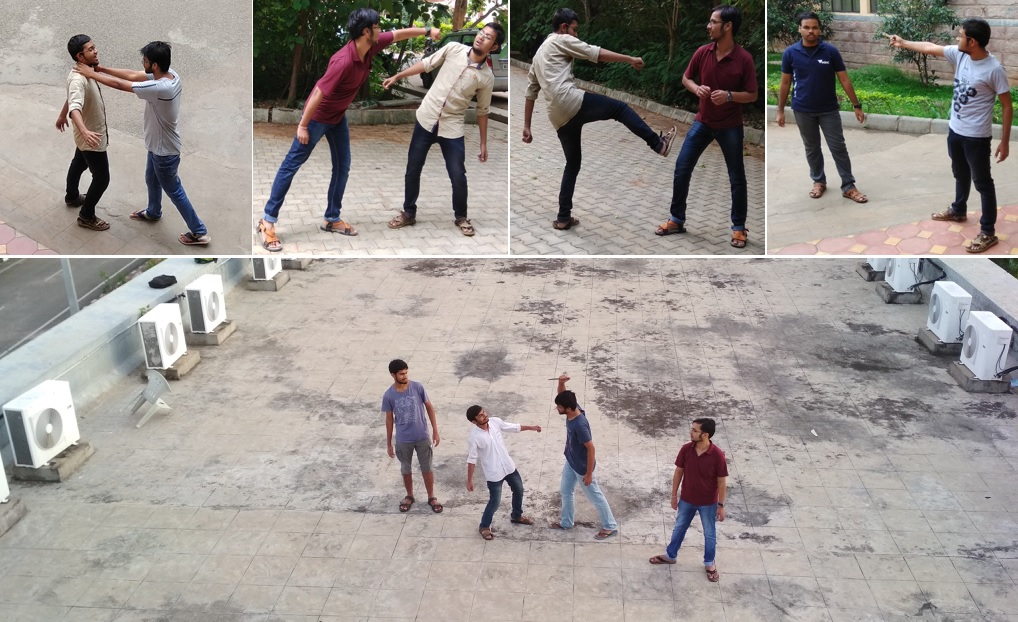}
	\end{center}
	\caption{\small{Illustration presents the violent activities from the introduced AVI dataset namely (clockwise from top) (i) Strangling,
			(ii) Punching, (iii) Kicking, (iv) Shooting and (v) Stabbing. The image of shooting activity involves multiple people in the same frame.}}
	\label{fig:long}
	\label{fig:onecol}
\end{figure}

\begin{figure}[t]
	\begin{center}
		\includegraphics[scale = 0.35]{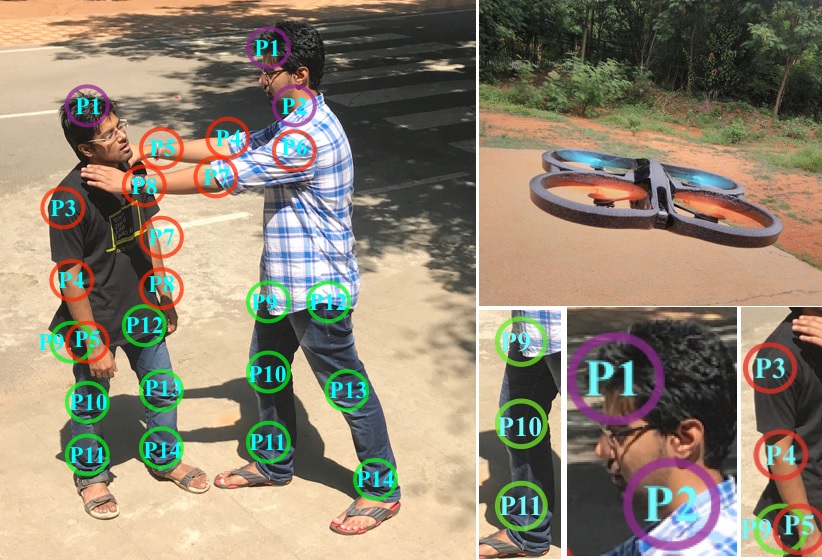}
	\end{center}
	\caption{\small{The figure (left) illustrates the 14 body key-points annotated on the human body. The description of the human body points is as Facial Region (Purple): P1-Head, P2- Neck; Arms Region (Red): P3- Right shoulder, P4- Right Elbow, P5- Right Wrist, P6- Left Shoulder, P7- Left Elbow, P8- Left Wrist; Legs Region (Green): P9-Right Hip, P10- Right Knee, P11-Right Ankle, P12- Left Hip, P13- Left Knee, P14- Left Ankle. The figure (right) shows the Parrot AR Drone used to capture the images in the dataset and close-ups of few annotated keypoints.}}
	\label{fig:long}
	\label{fig:onecol}
\end{figure}

\item \textbf{\textit{ScatterNet Hybrid Deep Network}}: The proposed SHDL network for pose estimation is composed of a hand-crafted ScatterNet front-end and a supervised learning based back-end formed of the modified coarse-to-fine deep regression network~\cite{belagiannis2015robust}, referred from now as the regression network (RN). The SHDL network is constructed by replacing the first convolutional, relu and pooling layers of the coarse-to-fine deep regression network~\cite{belagiannis2015robust} with the hand-crafted parametric log ScatterNet~\cite{singh}. This accelerates the learning of the regression network (RN) as the ScatterNet front-end extracts invariant (translation, rotation, and scale)~\cite{sifre2013} edge features which can be directly used to learn more complex patterns from the start of learning.  The invariant edge features can be beneficial for this application as the humans can appear with these variations in the aerial images. 

\item \textbf{\textit{Rapid Training with Structural Priors}}: Training of the SHDL network can be slow as it requires the optimization of several hyperparameters. The training is shown to accelerate by initializing the CNN layer filters of the regression network with structural priors learned (unsupervised) using the PCANet~\cite{pcanet} framework (Fig. 3). The initialization with priors also reduces the need for sizeable labeled training datasets for effective training which is especially advantageous for this task or other applications~\cite{tsshdl,jain} as it can be expensive and time-consuming to generate keypoint annotations.

\item \textbf{\textit{Real-time Identification}}: The proposed system performs the computation and memory demanding SHDL network processes along with the activity classification technique on the cloud while keeping short-term navigation onboard. This allows the system to identify violent individuals in real-time which is an improvement over the previous work of Surya et al.~\cite{penmetsa2014autonomous}. 

\item \textbf{\textit{Aerial Violent Individual (AVI) Dataset}}: The paper presents the  Aerial Violent Individual (AVI) dataset of 2000 annotated images (10863 total individuals) of 5124 individuals engaged in violent activities. The AVI dataset contains images with humans recorded at different variations of scale, position, illumination, blurriness, etc. This dataset may encourage researchers interested in using deep learning for aerial surveillance applications.
\end{itemize}

The proposed Drone Surveillance System (DSS) is used to identify the individuals engaged in violent activities from aerial images. The pose estimation and activity classification performance of the system is compared with the state-of-the-art techniques.

The paper is divided into the following sections. Section 2 presents the introduced AVI dataset while Section 3 introduces the proposed DSS system. Section 4 details the experimental results and Section 5 concludes this research.
\section{Aerial Violent Individual (AVI) Dataset}
This research proposes an annotated Aerial Violent Individual (AVI) dataset which is used by the proposed SHDL network to learn pose estimation. The dataset is composed of 2000 images where each image contains two and ten humans. The complete datasets consist of 10863 humans with 5124 (48\%) engaged in one or more of the five violent activities of  (1) Punching, (2) Stabbing, (3) Shooting, (4) Kicking, and (5) Strangling as shown in Fig. 1. Each human in the aerial image frame is annotated with 14 key-points which are utilized by the proposed network as labels for learning pose estimation as shown in Fig. 2. These activities are performed by 25 subjects between the ages of 18-25 years. These images are recorded from the parrot drone at four heights of 2m, 4m, 6m and 8m (m: meters). 

The violent individual identification task from these aerial images is an extremely challenging problem as these images can be affected by illumination changes, shadows, poor resolution, and blurring. In addition to these variations, the humans can appear at different locations, orientations, and scales. The proposed dataset includes images with the above-detailed variations as these can significantly alter the appearance of the humans and affect the performance of the surveillance systems. The SHDL network, when trained on the AVI dataset with these variations, can learn to recognize humans despite these variations. 

\section{Drone Surveillance System}
This section presents the Drone Surveillance System (DSS) for the identification of individuals engaging in violent activities. The system first uses the feature pyramid network (FPN)~\cite{hd} to detect humans from the images recorded by the drone. The proposed ScatterNet Hybrid Deep Learning (SHDL) Network is then used to estimate the pose of each detected human. Finally, the orientations between the limbs of the estimated pose are used to identify the violent individuals. The system uses cloud computation to achieve the identification in real-time. Each part of the Drone Surveillance System (DSS) is explained in the following sub-sections.

\begin{figure*}
	\begin{center}
		\includegraphics[width=1\linewidth]{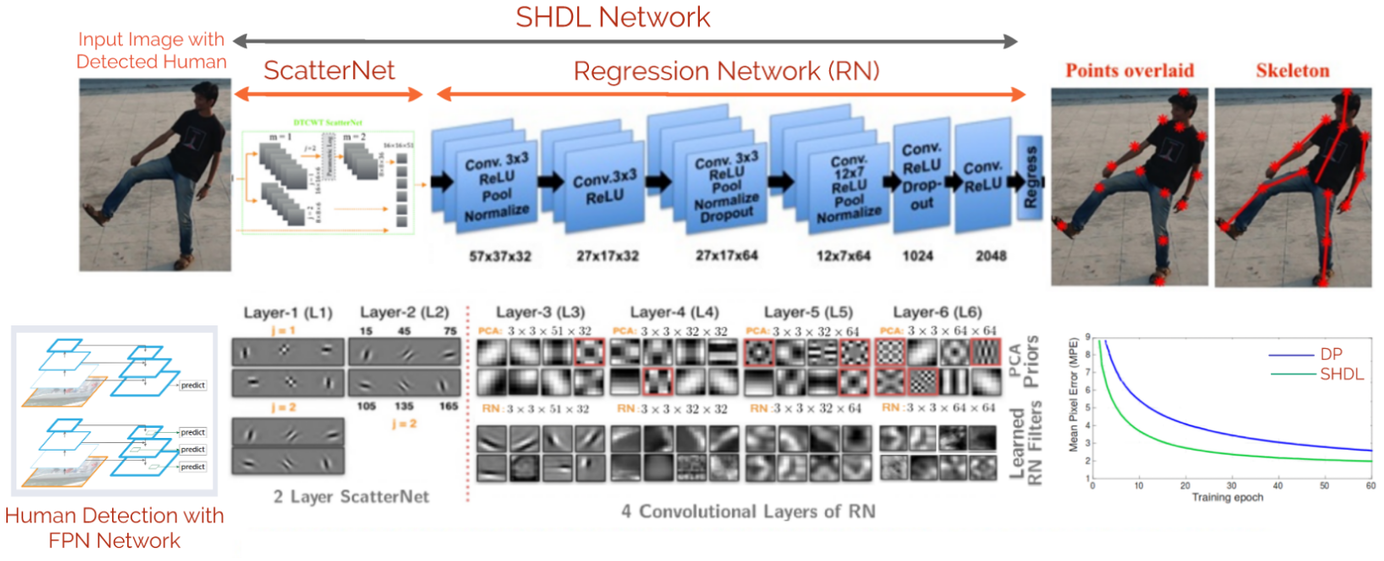}
	\end{center}
	\caption{\small{Illustration presents the \{textit{human pose estimation pipeline} that can be used to detect violent individuals in public areas or large gatherings. The DSS framework uses the image recorded by the drone first to discover the humans within the image using the FPN network~\cite{hd}. The image regions containing the humans are given as input to the proposed SHDL network to detect 14 key-points on the body for pose estimation. The proposed SHDL network uses the ScatterNet (front-end) to extract hand-crafted features from the input region at L0, L1, and L2 using DTCWT filters at two scales and six fixed orientations. The handcrafted features extracted from the three layers are concatenated and given as input to the 4 Convolutional layers of the Regression Network (RN) (L3, L4, L5, L6) (back-end) with 32, 32, 64 and 64 filters. Each RN convolutional layer is initialized with the PCA based structural priors with the same number of filters. PCA layers can learn the undesired checkerboard filters (shown in red) which are avoided and not used as the prior for the Regression Network. To detect and remove the checkerboard filters from the learned filter set; we used the method defined in ~\cite{geiger2012}. The ScatterNets and Structural priors have shown to improve the training of the proposed SHDL network as compared to the original coarse-to-fine regression network~\cite{belagiannis2015robust} (which was modified to obtain the SHDL) as shown from the convergence graph. The 14 key-points detected on the human are connected to construct the skeleton structure. The hand-crafted filters for the ScatterNet, learned structural PCA priors and the learned filters of the regression network (RN) are shown. }}
	\label{fig:short}
\end{figure*}

\subsection{Human Detection}
The DSS system makes uses of the feature pyramid network (FPN)~\cite{hd} to detect humans quickly from the images recorded by the drone. The FPN network detects the humans by leveraging the pyramidal shape of a ConvNet’s feature hierarchy while creating a feature pyramid that has strong semantics at all scales. The result is a feature pyramid that has rich semantics at all levels and is built quickly from a single input image scale. 

\subsection{ScatterNet Hybrid Deep Learning Network}
This section details the proposed ScatterNet Hybrid Deep Learning (SHDL) network, inspired from Singh et al.'s work in~\cite{eff,shdl2017,tsshdl, GSHDL}, composed by combining the hand-crafted (front-end) two-layer parametric log ScatterNet~\cite{singh} with the regression network (RN) (back-end)  shown in Fig. 3. The ScatterNet accelerates the learning of the SHDL network by extracting invariant edge-based features which allow the SHDL network to learn complex features from the start of the learning~\cite{eff}. The regression network also uses structural priors to expedite the training as well as reduce the dependence on the annotated datasets. The ScatterNet (front-end) and regression network (RN) (back-end) parts of the proposed SHDL network are presented below.

\vspace{0.5mm}
\textbf{\textit{ScatterNet (front-end)}}: The parametric log based DTCWT ScatterNet~\cite{singh} is an improved numerous version of the hand-crafted multi-layer Scattering Networks~\cite{Jbruna2013,ima,eccv} proposed over the years. The parametric log ScatterNet extracts relatively symmetric translation invariant representations using the \textit{dual-tree complex wavelet transform} (DTCWT)~\cite{Kingsbury1998} and parametric log transformation layer. The ScatterNet features are denser over scale as they are extracted from multi-resolution images at 1.5 times and twice the size of the input image. Below we present the formulation of the parametric DTCWT ScatterNet for a single input image which may then be applied to each of the multi-resolution images.

The parametric log ScatterNet is a hand-crafted two-layer network which extracts translation invariant feature representation from an input image or signal. The invariant features are obtained at the first layer by filtering the input signal $x$ with dual-tree complex wavelets (better than cosine transforms~\cite{wavdct}) $ \psi_{j,r }$ at different scales ($j$) and six pre-defined orientations ($r$) fixed to $15^\circ, 45^\circ, 75^\circ, 105^\circ, 135^\circ$ and $165^\circ$. To build a more translation invariant representation, a point-wise $L_{2}$ non-linearity (complex modulus) is applied to the real and imaginary part of the filtered signal:
\begin{equation}
U[\lambda_{m = 1}] = |x\star \psi_{\lambda_{1} }| = \sqrt{|x\star \psi_{\lambda_{1} }^{a}|^2 + |x\star \psi_{\lambda_{1} }^{b}|^2} 
\end{equation}
The parametric log transformation layer is then applied to all the oriented representations extracted at the first scale $j=1$ with a parameter $k_{j=1}$, to reduce the effect of outliers by introducing relative symmetry of pdf~\cite{singh}, as shown below: 
\begin{equation}
U1[j] = \log(U[j] + k_{j}), \quad U[j] = |x\star \psi_{j}|, 
\end{equation}
Next, a local average is computed on the envelope $|U1[\lambda_{m = 1}]|$ that aggregates the coefficients to build the desired translation-invariant representation: 
\begin{equation}
S_{1}[\lambda_{m = 1}] = |U1[\lambda_{m = 1}]| \star \phi_{2^J}
\end{equation}
The high frequency components lost due to smoothing are retrieved by cascaded wavelet filtering performed at the second layer. Translation invarinace is introduced in these features by applying the L2 non-linearity with averaing as explained above for the first layer~\cite{singh}.

The scattering coefficients at L0, L1, and L2 are:
\vspace{-0.3em}
\begin{equation}
S = \begin{pmatrix}
x \star \phi_{2^J},
S_{1}[\lambda_{m = 1}],S_{2}[\lambda_{m = 1},\lambda_{m = 2}] \star \phi_{2^J}
\end{pmatrix}
\end{equation} 

The rotation and scale invariance are next obtained by filtering jointly across the position ($u$), rotation ($\theta$) and scale($j$) variables as detailed in~\cite{sifre2013}. 

The features extracted from each multi-resolution at L0, L1, and L2 are concatenated and given as input to the regression network (RN), to learn high-level features for human pose estimation. The ScatterNet features help the proposed SHDL to converge faster as the convolutional layers of the regression network can learn more complex patterns from the start of learning as it is not necessary to wait for the first layer to learn invariant edges as the ScatterNet already extracts them.

\textbf{\textit{Pose Estimation with Structural Priors (back-end):}}
The invariant ScatterNet features are used by the regression network (RN) of the SHDL network to learn pose estimation from the AVI dataset. The regression network was constructed by removing the first convolutional, relu, pooling, and normalization layers of the coarse-to-fine deep regression network~\cite{belagiannis2015robust}. The regression network (RN) of the SHDL is composed of four convolutional (L3 to L6 layers), two fully connected, normalization, and max-pooling layers as shown in Fig. 3. 

The training objective is to estimate the optimal weights of the filters in the convolutional layers using the AVI training dataset $D = (S; Y)$, which minimizes the Tukey's biweight loss function~\cite{belagiannis2015robust} of the network. Here $S$ are the ScatterNet features extracted from the input image ($X$) while $Y$ is a 28  element vector of ($x,y$) corresponding to the 14 key-points annotated on the human body as shown in Fig. 2. The network is optimized using backpropagation with stochastic gradient descent. Dropout is utilized to avoid overfitting. Tukey's biweight loss function is very efficient as it suppresses the influence of outliers during backpropagation by reducing the magnitude of the gradient close to zero~\cite{belagiannis2015robust}. 

\textbf{\textit{Structural Priors}}: Each convolutional layer (L3 to L6) of the regression network (RN) of the SHDL network is initialized with structural priors to accelerate the training. The Structural priors are obtained for each layer using the PCANet~\cite{pcanet} framework that learns a family of orthonormal filters by minimizing the following reconstruction error:
\begin{figure}[t]
	\begin{center}
		\includegraphics[scale = 0.5]{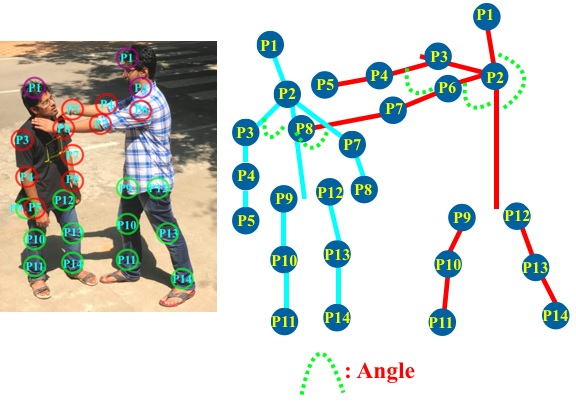}
	\end{center}
	\caption{
		The Illustration shows the skeleton corresponding to the humans in an image. The angles (shown in green for few limbs) between the various limbs in this structure are used by the SVM to recognize the humans engaged in violent activities.}
	\label{fig:long}
	\label{fig:onecol}
\end{figure}
\begin{equation}
\min_{V \epsilon \ R^{z_{1}z_{2}\times K} } \left \|X-VV^TX  \right \|_{F}^2,\ s.t.\  VV^T = I_{K}
\end{equation} 
Where $X$ are patches sampled from $N$ training features, $I_K$ is an identity matrix of size $K \times K$. The solution of Eq. 5 in its simplified form represents $K$ leading principal eigenvectors of $XX^T$ obtained using Eigen decomposition.

The structural priors for layer 3 (L3) are learned on the ScatterNet features, layer 4 (L4) on layer 3 outputs, layer 5 (L5) on layer 4 outputs and so on. The structural priors for L3 to L6 layers learn filters that respond to a hierarchy of features, similar to the features learned by the CNN's. These learned priors are used to initialize each convolutional layer resulting in accelerated training as shown in Fig. 3 (Graph). Since it is swift to determine the structural priors, the whole process is much quicker than training CNN's with random weight initialization. The PCA framework may learn undesired checkerboard filters. To detect the checker-board filters from the learned filter set, we use the method defined in~\cite{geiger2012}. These checkerboard filters are avoided as filter priors. 

\subsection{\textbf{\textit{Violent Individual Classification}}}
The 14 key-points identified by the SHDL network are connected to form a skeleton structure as shown in Fig. 3. The orientations between the limbs of the skeleton structure are derived as shown in Fig. 4.  A support vector machine (SVM) is trained on a vector of these orientations for six classes (five violent activities and one neutral activity) to perform multi-class classification. During test time, the orientations between the limbs of the skeleton are given as input to the SVM which classifies the humans as either neural or assigns the most likely violent activity label.

	


\begin{figure*}
	\begin{center}
		\includegraphics[width=1\linewidth]{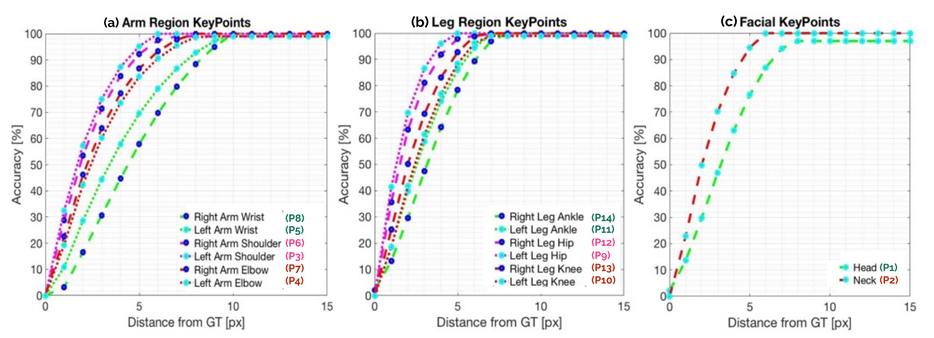}
	\end{center}
	\caption{
		Illustration shows the pose estimation performance via the detection of key-points for the (a) arms region, which constitutes the wrist, shoulder and elbow, (b) legs region, which includes ankle, knee, and hip, and, (c) facial regions with the head and neck.}
	\label{fig:short}
\end{figure*}

\subsection{Drone Image Acquisition and Cloud Processing}
The images that form the AVI dataset, presented in Section 2 are recorded using a Parrot AR Drone. The AR Drone 2.0 consists of two cameras, an Inertial Measurement Unit (IMU) including a 3-axis accelerometer, 3-axis gyroscope and 3-axis magnetometer, and ultrasound and pressure-based altitude sensors. It features a 1 GHz ARM Cortex-A8 as the CPU and runs a Linux operating system.  The front-facing camera has a resolution of 1280$\times$720 at 30fps with a diagonal field of view of $92^\circ$ while the downward facing camera is of the lower resolution of 320$\times$240 at 60fps with a diagonal field of view of $64^\circ$. We use the front-facing camera to record the images due to its higher resolution. The downward facing camera estimates the parameters determining the state of the drone such as roll, pitch, yaw, and altitude using the sensors onboard to measure the horizontal velocity. The horizontal velocity calculation is based on an optical flow-based feature as detailed in  \cite{bristeau2011navigation}. All the sensor measurements are updated at the 200Hz rate. 

The images recorded by the drone are transferred to the Amazon cloud to achieve real-time identification.  The slow and memory intensive computations of the SHDL network are processed on the Amazon cloud while keeping short-term navigation onboard. Cloud computing has given the flexibility of using unlimited computational resources (including GPUs) which provides an edge with for applications requiring vast amounts of computational power periodically \cite{goldberg2013cloud}.

\section{Experimental Results}
This section presents the training details and the performance of the Drone Surveillance System (DSS) for the identification of violent individuals on the AVI dataset. The DSS system uses the FPN network~\cite{hd} first to detect the humans, the SHDL network for human pose estimation, and then the orientations of the limbs of the estimated pose are used to identify the violent individuals. The next sections detail the performance of each part of the DSS system. The classification performance is also compared with the state-of-the-art technique proposed by Surya et al.~\cite{penmetsa2014autonomous}, used to identify persons of interest from aerial images.

\subsection{\textbf{\textit{Human Detector}}}
The FPN network~\cite{hd} pre-trained on the 80 category COCO detection dataset is used to detect the humans recorded by the drone in the AVI dataset. The FPN network was able to detect 10558 humans out of the 10863 humans, with an accuracy of 97.2\%.

\begin{figure*}
	\begin{center}
		\includegraphics[width=1\linewidth]{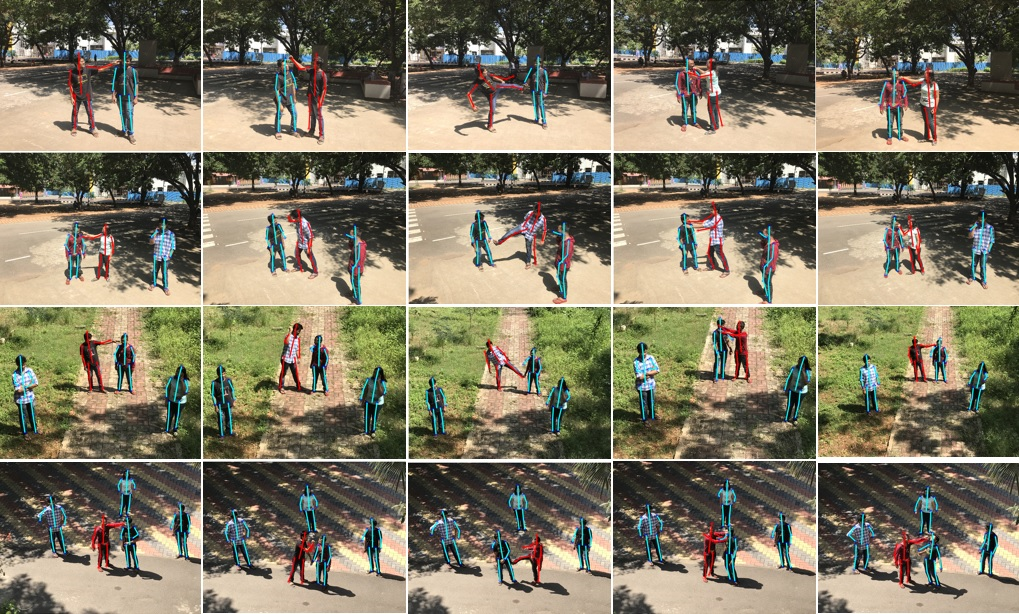}
	\end{center}
	\caption{\small{The figure shows the performance of the Drone Surveillance System (DSS) on aerial images with only one violent individual, recorded using the AR parrot drone at four different heights of 2m (Row 1), 4m (Row 2), 6m (Row 3), and 8m (Row 4) (m: meters). The illustration also shows the individual engaged in different violent activities namely: Shooting (Column 1), Stabbing (Column 2), Kicking (Column 3), Strangling (Column 4) and Punching (Column 5). The violent individual detected by the DSS framework is shown in red while the neutral human is shown in cyan color.  The estimated pose is also shown on top each detected human.}}
	\label{fig:short}
\end{figure*}

\subsection{\textbf{\textit{SHDL Parameters and Training}}}
The image regions detected by the FPN network are resized to 120 $\times$ 80 and normalized by subtracting the image regions mean and dividing by its standard deviation. 

\textbf{\textit{ScatterNet}}: The resultant image region is given as input to the ScatterNet (SHDL front-end)  which extracts invariant edge representations at L0, L1, and L2 using DTCWT filters at 2 scales, and 6 fixed orientations.

\textbf{\textit{Regression Network with Structural Priors}}: The regression network (SHDL back-end) with four convolutional layers (L3-L6) is trained on the concatenated ScatterNet features (L0, L1, and L2) extracted from the 10558 image regions (detected by FPN network, Section 4.1). The network was trained on randomly selected 6334 image regions (60\%), validated against 2111 image regions (20\%) and tested on the remaining 2113 image regions (20\%). The network parameters are as follows: The base learning rate is $10^{-5}$, which we decrease to $10^{-6}$ after 20 iterations, the dropout is 0.5, the batch size is 20, and the total number of iterations (epochs) is 90. The filters of the convolutional layers are initialized with structural priors which are shown to accelerate the training as compared to the DeepPose network~\cite{toshev2014deeppose} as detailed in Section 3.2 and illustrated from the convergence graph in Fig. 3.

\subsection{\textbf{\textit{Key-Point Detection Performance}}}
The pose estimation performance of the SHDL network is evaluated by comparing the coordinates of the detected 14 key-points with their ground truth values on the annotated dataset. The key-point is deemed correctly located if it is within a set distance of $d$ pixels from a marked key-point in the ground truth, as shown in Fig. 5 via the accuracy vs. distance graphs, for different regions of the body.

The key-points detection analysis for the arms, legs, and facial, region is presented below. 

\textbf{\textit{Arms Region}}: The arm region constitutes six points namely: wrist key-points (P5 and P8), shoulder key-points (P3 and P6), and elbow key-points(P4 and P7), as shown in Fig. 2. Fig. 5(a) indicates that the SHDL network can detect the wrist region key-points with an accuracy of around 60\%, for a pixel distance of d=5. The detection accuracy is much higher for the elbow and shoulder region at roughly 85\% and 95\% respectively, for the same pixel distance (d=5).

\textbf{\textit{Legs Region}}: The leg region constitutes six key-points, namely: hip key-points (P9, P12), knee key-points (P10, P13), and ankle key-points (P11, P14), as shown in Fig. 2. Fig. 5(b) indicates that the SHDL network detects hip key-points with almost 100\% for a pixel distance of d=5. The detection accuracy is between 85\% and 90\% for the knee key-points while the detection rate falls to around 85\% for the ankle key-points.

\textbf{\textit{Facial Region}}: The facial region constitutes two points, one the head (P1) and the other on the neck (P2), as shown in Fig. 2. The algorithm detects the neck key-point (P2) more accurately as compared the head key-point (P1) with an accuracy of around 95\% as opposed to roughly 77\% accuracy, for a pixel distance of d=5, as shown in Fig. 5(c).

The human pose estimation performance of the SHDL network on the Aerial Violent Individual (AVI) dataset is presented in Table 1. As observed from the Table, the SHDL network estimates the human pose based on the 14 key-points at d = 5 pixel distance from the ground-truth, with 87.6\% accuracy. 

\begin{table}[!h]%
	\centering
	\begin{tabular}{c|cccc}
		\hline
		\multicolumn{1}{c}{Dataset} & \multicolumn{4}{c}{Deep Learning Networks}   \\ 
		\hline
		&  SHDL & CN  & CNE & SpatialNet   \\
		\cline{2-4} \hline
		\small{AVI} & \textbf{87.6} & 79.6 & 80.1 & 83.4 \\ 
	\end{tabular}
	\newline
	\caption{{Comparison of the human pose estimation performance of SHDL network with Coordinate network (CN) ~\cite{cne}, Coordinate extended network (CNE)~\cite{cne} ~\cite{pfister2015flowing} and Spatial network~\cite{pfister2015flowing} based on the detection of the 14 key-points. The evaluation is presented on the AVI dataset for maximum 5 pixels allowed distance (d=5) from the annotated ground truth.}}
\end{table}

The human pose estimation performance of the SHDL network is also compared with several state-of-the-art pose estimation methods such as CoordinateNet (CN)~\cite{cne}, CoordinateNet extended(CNE)~\cite{cne}, and SpatialNet ~\cite{pfister2015flowing}. The proposed SHDL network outperforms them by a decent margin.

\begin{figure*}
	\begin{center}
		\includegraphics[width=1\linewidth]{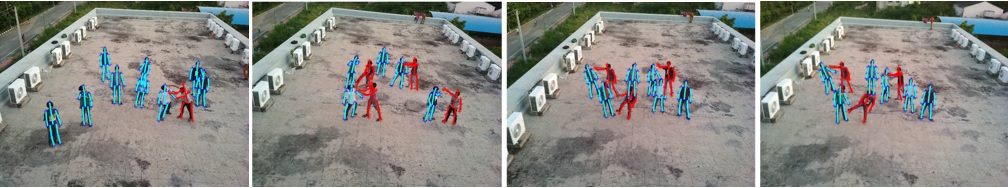}
	\end{center}
	\caption{The figure shows the performance of the Drone Surveillance System (DSS) on aerial images with multiple humans engaging together in different violent activities. The violent individuals are highlighted in red color and neutral human in cyan color. }
	\label{fig:short}
\end{figure*}

\subsection{\textbf{\textit{Violent Individuals Identification}}}
The detected key-points are connected to form a skeleton structure as shown in Fig. 3. The orientations between the limbs are concatenated as a vector. A support vector machine (SVM) with a Gaussian kernel is trained on the orientation vector for each class of violent activity and one neutral class for 6334 randomly selected human poses (60\%) to perform the multi-class classification. The SVM parameter (c) is selected as 14 while gamma parameter is set to 0.00002 using 5-fold cross-validation on the training set. The classification accuracy on the AVI dataset of each violent activity is presented for 4224 (40\%) human poses as shown in Table 2.

\begin{table}[!h]%
	\centering
	\scalebox{0.9}{
		\begin{tabular}{c|ccccc}
			\hline
			\multicolumn{1}{c}{\small{Dataset}} & \multicolumn{5}{c}{Violent Activities}   \\ 
			\hline
			&  \small{Punching}  & \small{Kicking} & \small{Strangling} & \small{Shooting} & \small{Stabbing} \\
			\cline{2-4} \hline
			\small{DSS} & 89 & 94 & 85 & 82 & 92 \\
			\small{\small{Surya}~\cite{penmetsa2014autonomous}} &  80 & 84 & 73 & 73 & 79\\ 
			
		\end{tabular}
	}
	\newline
	\caption{Table presents the classification accuracies(\%) for the violent activities on Aerial Violent Individual (AVI) dataset. }
\end{table}

The accuracy of the strangling and shooting activities are relatively lower due to their similarity as shown in Fig. 6.  

Next, the classification accuracy for varying number of human subjects engaged in a violent activity per image is shown in Table 3.

\begin{table}[!h]%
	\centering
	\scalebox{0.9}{
		\begin{tabular}{c|ccccc}
			\hline
			\multicolumn{1}{c}{\small{Dataset}} & \multicolumn{5}{c}{No. of Violent Individuals (Per Image)}   \\ 
			\hline
			&  \small{1}  & \small{2} & \small{3} & \small{4} & \small{5} \\
			\cline{2-4} \hline
			\small{DSS} & 94.1 & 90.6 & 88.3 & 87.8 & 84.0
			
		\end{tabular}
	}
	\newline
	\caption{The table presents the classification accuracies(\%) with the increase in individuals engaged in the violent activities in the aerial images taken the Aerial Violent Individual (AVI) dataset.}
\end{table}
The accuracy of the DSS system decreases with the increase in the number of humans in the aerial image. This can be due to the inability of the FPN network~\cite{hd} to locate all the humans or the incapability of the SHDL network to estimate the pose of the humans accurately. The incorrect pose can result in a wrong orientations vector which can lead the SVM to classify the activities incorrectly.   

The results presented in above table are encouraging as the system is more likely to encounter multiple people in an image frame. The DSS framework applied to images with the different number of people engaged in violent activities is shown in Fig. 7. 

The classification performance is also compared with the state-of-the-art technique which was developed to recognize the person of interest from aerial images~\cite{penmetsa2014autonomous} as shown in Table. 4. The proposed Drone Surveillance System (DSS) was able to outperform the method by more than 10\% on the AVI dataset. 

\begin{table}[!h]%
	\centering
	
	\begin{tabular}{c|c c }
		\hline
		\multicolumn{1}{c}{Dataset} & \multicolumn{2}{c}{Comparison}   \\ 
		\hline
		& DSS & state-of-the-art~\cite{penmetsa2014autonomous}  \\
		\cline{2-3} \hline
		\small{AVI} & \textbf{88.8} & 77.8 \\ 
	\end{tabular}
	\newline
	\caption{
		The table shows the comparison of the violent individual identification performance of the proposed system against the state-of-the-art technique~\cite{penmetsa2014autonomous}}
\end{table}

\subsection{\textbf{\textit{Runtime Performance}}}
The runtime performance of the DSS framework is computed on the cloud and consists of three parts: (i) detecting humans using the FPN network, (ii) human pose estimation using the SHDL network, and (iii) classification of the estimated pose. The deep learning framework was accelerated using the cuDNN framework and NVIDIA Tesla GPUs. The system detected the violent individuals at 5 fps per second to 16 fps for a maximum of ten and a minimum of two people, respectively, in the aerial image frame. The processing varies depending on the number of individuals within the image frame.

\section{Conclusions}
The paper proposed the real-time Drone Surveillance System (DSS) framework that can detect one or more individuals engaged in violent activities from aerial images. The framework first uses the FPN network to detect humans after which the proposed SHDL network is used to estimate the pose of the humans. The estimated poses are used by the SVM to identify violent individuals. 

The proposed SHDL network uses ScatterNet features with Structural priors to achieve accelerated training for relatively fewer labeled examples. The utilization of fewer labeled examples for pose estimation is beneficial for this application as it is expensive to collect annotated examples. The paper also introduced the Aerial Violent Individual (AVI) Dataset which can benefit other researcher aiming to use deep learning for aerial surveillance applications. The proposed DSS framework outperforms the state-of-the-art technique on the AVI dataset. This framework will be instrumental in detecting individuals engaged in violent activities in public areas or large gatherings.

{\small
	\bibliographystyle{ieee}
	\bibliography{egbib}

\begin{thebibliography}{10}\itemsep=-1pt

\bibitem{belagiannis2015robust}
V.~Belagiannis, C.~Rupprecht, G.~Carneiro, and N.~Navab.
\newblock Robust optimization for deep regression.
\newblock In {\em Proceedings of the IEEE International Conference on Computer
  Vision}, pages 2830--2838, 2015.

\bibitem{bristeau2011navigation}
P.-J. Bristeau, F.~Callou, D.~Vissi{\`e}re, N.~Petit, et~al.
\newblock The navigation and control technology inside the ar. drone micro uav.
\newblock In {\em 18th IFAC World Congress}, 2011.

\bibitem{Jbruna2013}
J.~Bruna and S.~Mallat.
\newblock Invariant scattering convolution networks.
\newblock {\em IEEE transactions on pattern analysis and machine intelligence},
  35(8):1872--1886, 2013.

\bibitem{pcanet}
T.-H. Chan, K.~Jia, S.~Gao, J.~Lu, Z.~Zeng, and Y.~Ma.
\newblock Pcanet: A simple deep learning baseline for image classification?
\newblock {\em IEEE Transactions on Image Processing}, 2015.

\bibitem{chuang2009carried}
C.-H. Chuang, J.-W. Hsieh, L.-W. Tsai, S.-Y. Chen, and K.-C. Fan.
\newblock Carried object detection using ratio histogram and its application to
  suspicious event analysis.
\newblock {\em IEEE transactions on circuits and systems for video technology},
  2009.

\bibitem{felzenszwalb2005pictorial}
P.~F. Felzenszwalb and D.~P. Huttenlocher.
\newblock Pictorial structures for object recognition.
\newblock {\em International journal of computer vision}, 61(1):55--79, 2005.

\bibitem{geiger2012}
A.~Geiger, F.~Moosmann, {\"O}.~Car, and B.~Schuster.
\newblock Automatic camera and range sensor calibration using a single shot.
\newblock In {\em Robotics and Automation (ICRA), 2012 IEEE International
  Conference on}, pages 3936--3943, 2012.

\bibitem{goldberg2013cloud}
K.~Goldberg and B.~Kehoe.
\newblock Cloud robotics and automation: A survey of related work.
\newblock {\em University of California, Berkeley, Tech. Rep. UCB/EECS-2013-5},
  2013.

\bibitem{goya2009method}
K.~Goya, X.~Zhang, K.~Kitayama, and I.~Nagayama.
\newblock A method for automatic detection of crimes for public security by
  using motion analysis.
\newblock In {\em International Conference on Intelligent Information Hiding
  and Multimedia Signal Processing}, 2009.

\bibitem{jain}
S.~Jain, S.~Gupta, and A.~Singh.
\newblock A novel method to improve model fitting for stock market prediction.
\newblock {\em International Journal of Research in Business and Technology},
  3(1):78--83.

\bibitem{wavdct}
V.~Jeengar, S.~Omkar, A.~Singh, M.~K. Yadav, and S.~Keshri.
\newblock A review comparison of wavelet and cosine image transforms.
\newblock {\em International Journal of Image, Graphics and Signal Processing},
  4(11):16, 2012.

\bibitem{Kingsbury1998}
N.~G. Kingsbury.
\newblock The dual-tree complex wavelet transform: a new technique for shift
  invariance and directional filters.
\newblock In {\em Proc. 8th IEEE DSP workshop}, volume~8, 1998.

\bibitem{lewis2010cctv}
P.~Lewis.
\newblock Cctv in the sky: police plan to use military-style spy drones.
\newblock {\em The Guardian}, 23, 2010.

\bibitem{li20143d}
S.~Li and A.~B. Chan.
\newblock 3d human pose estimation from monocular images with deep
  convolutional neural network.
\newblock In {\em Asian Conference on Computer Vision}, pages 332--347, 2014.

\bibitem{li2010abandoned}
X.~Li, C.~Zhang, and D.~Zhang.
\newblock Abandoned objects detection using double illumination invariant
  foreground masks.
\newblock In {\em Pattern Recognition (ICPR), 2010 20th International
  Conference on}, pages 436--439, 2010.

\bibitem{hd}
T.-Y. Lin, P.~Doll{\'a}r, R.~Girshick, K.~He, B.~Hariharan, and S.~Belongie.
\newblock Feature pyramid networks for object detection.
\newblock 2017.

\bibitem{eccv}
S.~Nadella, A.~Singh, and S.~Omkar.
\newblock Aerial scene understanding using deep wavelet scattering network and
  conditional random field.
\newblock In {\em European Conference on Computer Vision}, pages 205--214,
  2016.

\bibitem{padgett2009drones}
T.~Padgett.
\newblock Drones join the war against drugs.
\newblock {\em Time Magazine, June}, 2009.

\bibitem{penmetsa2014autonomous}
S.~Penmetsa, F.~Minhuj, A.~Singh, and S.~Omkar.
\newblock Autonomous uav for suspicious action detection using pictorial human
  pose estimation and classification.
\newblock {\em ELCVIA: electronic letters on computer vision and image
  analysis}, 13(1):18--32, 2014.

\bibitem{cne}
T.~Pfister.
\newblock Advancing human pose and gesture recognition.
\newblock In {\em University of Oxford}, 2015.

\bibitem{pfister2015flowing}
T.~Pfister, J.~Charles, and A.~Zisserman.
\newblock Flowing convnets for human pose estimation in videos.
\newblock In {\em IEEE International Conference on Computer Vision}, 2015.

\bibitem{pfister2014deep}
T.~Pfister, K.~Simonyan, J.~Charles, and A.~Zisserman.
\newblock Deep convolutional neural networks for efficient pose estimation in
  gesture videos.
\newblock In {\em Asian Conference on Computer Vision}, pages 538--552, 2014.

\bibitem{seebamrungsat2014fire}
J.~Seebamrungsat, S.~Praising, and P.~Riyamongkol.
\newblock Fire detection in the buildings using image processing.
\newblock In {\em Student Project Conference (ICT-ISPC), 2014 Third ICT
  International}, pages 95--98, 2014.

\bibitem{sifre2013}
L.~Sifre and S.~Mallat.
\newblock Rotation, scaling and deformation invariant scattering for texture
  discrimination.
\newblock In {\em Computer Vision and Pattern Recognition (CVPR), 2013 IEEE
  Conference on}, pages 1233--1240, 2013.

\bibitem{tsshdl}
A.~Singh, D.~Hazarika, and A.~Bhattacharya.
\newblock Texture and structure incorporated scatternet hybrid deep learning
  network (ts-shdl) for brain matter segmentation.
\newblock {\em International Conference on Computer Vision Workshop}, 2017.

\bibitem{ima}
A.~Singh and N.~Kingsbury.
\newblock Multi-resolution dual-tree wavelet scattering network for signal
  classification.
\newblock {\em International Conference on Mathematics in Signal Processing},
  2016.

\bibitem{singh}
A.~Singh and N.~Kingsbury.
\newblock Dual-tree wavelet scattering network with parametric log
  transformation for object classification.
\newblock In {\em International Conference on Acoustics, Speech and Signal
  Processing (ICASSP)}, 2017.

\bibitem{eff}
A.~Singh and N.~Kingsbury.
\newblock Efficient convolutional network learning using parametric log based
  dual-tree wavelet scatternet.
\newblock {\em IEEE International Conference on Computer Vision Workshop},
  2017.

\bibitem{shdl2017}
A.~Singh and N.~Kingsbury.
\newblock Scatternet hybrid deep learning (shdl) network for object
  classification.
\newblock {\em International Workshop on Machine Learning for Signal
  Processing}, 2017.

\bibitem{GSHDL}
A.~Singh and N.~Kingsbury.
\newblock Generative scatternet hybrid deep learning (g-shdl) network with
  structural priors for semantic image segmentation.
\newblock {\em IEEE International Conference on Acoustics, Speech and Signal
  Processing}, 2018.

\bibitem{toshev2014deeppose}
A.~Toshev and C.~Szegedy.
\newblock Deeppose: Human pose estimation via deep neural networks.
\newblock In {\em Proceedings of the IEEE Conference on Computer Vision and
  Pattern Recognition}, pages 1653--1660, 2014.

\bibitem{walters2010ucav}
W.~Walters and J.~Weber.
\newblock Ucav surveillance, high-tech masculinities and oriental others.
\newblock {\em presentation to A Global Surveillance Society}, 2010.

\end{thebibliography}
}

\end{document}